\documentclass[letterpaper, 10 pt, conference]{ieeeconf}  

\IEEEoverridecommandlockouts                              

\usepackage{graphics} 
\usepackage{epsfig} 
\usepackage{mathptmx} 
\usepackage{times} 
\usepackage{amsmath} 
\usepackage{amssymb}  
\usepackage{bm}
\usepackage{algorithm}
\usepackage{multirow}
\usepackage{verbatim}
\usepackage{graphicx}
\usepackage{hyperref}
\usepackage{booktabs} 

\title{\LARGE \bf
A Proposed Set of Communicative Gestures for Human Robot Interaction and an RGB Image-based Gesture Recognizer Implemented in ROS
}

\author{Jia Chuan A. Tan$^{1}$, Wesley P. Chan$^{1}$, Nicole L. Robinson$^{1}$, Elizabeth A. Croft$^{1}$, Dana Kuli\'c$^{1}$
\thanks{$^{1}$Monash University}
} 
\begin{document}

\maketitle
\thispagestyle{empty}
\pagestyle{empty}

\begin{abstract}
We propose a set of communicative gestures and develop a gesture recognition system with the aim of facilitating more intuitive Human-Robot Interaction (HRI) through gestures. First, we propose a set of commands commonly used for human-robot interaction. Next, an online user study with 190 participants was performed to investigate if there was an agreed set of gestures that people intuitively use to communicate the given commands to robots when no guidance or training were given. As we found large variations among the gestures exist between participants, we then proposed a set of gestures for the proposed commands to be used as a common foundation for robot interaction. We collected $\sim7500$ video demonstrations of the proposed gestures and trained a gesture recognition model, adapting 3D Convolutional Neural Networks (CNN) as the classifier, with a final accuracy of $84.1\pm{2.4}\%$. The resulting model was capable of training successfully with a relatively small amount of training data. We integrated the gesture recognition model into the ROS framework and report details for a demonstrated use case, where a person commands a robot to perform a pick and place task using the proposed set. This integrated ROS gesture recognition system is made available for use, and built with the intention to allow for new adaptations depending on robot model and use case scenarios, for novel user applications. 

\end{abstract}

\section{INTRODUCTION}
        \label{sec:introduction}
        Facilitating intuitive Human-Robot Interaction is important for increasing robot use in both domestic and industrial settings~\cite{noauthor_1_nodate}. However, current interfaces such as teach-pendants do not allow for programming during interaction with a robot~\cite{villani_survey_2018}. These approaches are commonly used in relatively static or highly engineered, controlled environments, with robots isolated from their operator. As a result, there is limited interactivity between pre-programmed robots and workers in current systems~\cite{haddadi_analysis_2013}. 

A more intuitive, interactive form of instruction could facilitate more fluid collaboration, engaged interaction, and improved safety when working with a robot. Newer methods include gesture recognition, which involves humans using their hands or body to provide gesture-based signals to robots. Gesture recognition is used during human-human communication, and this can be translated into human-robot interaction~\cite{6483609,HariRiitta2009BBoH}. 

Gestural communication removes some challenges seen with speech recognition, such as language complexity, different accents, adaptive phrases and loud environments~\cite{doi:10.1089/dst.2012.0010,6926324}. Gesture recognition can allow humans operators to communicate with a robot and replan HRI tasks in real time~\cite{chen_online_2019, kim_vision-based_2013}. However, to date, the selection of the gesture set for Human-Robot Communication has not been systematically explored. Gesture communication sets should be easy to use, gestures should be understood by other humans who are in the environment, quick to adapt to multiple robot platforms, and should be easily adapted in a quick and efficient manner if people using the system need to reprogram or update the current list of communicative gestures or command sets.  

\begin{figure} [t]
    \centering
    \includegraphics[width=0.48\textwidth]{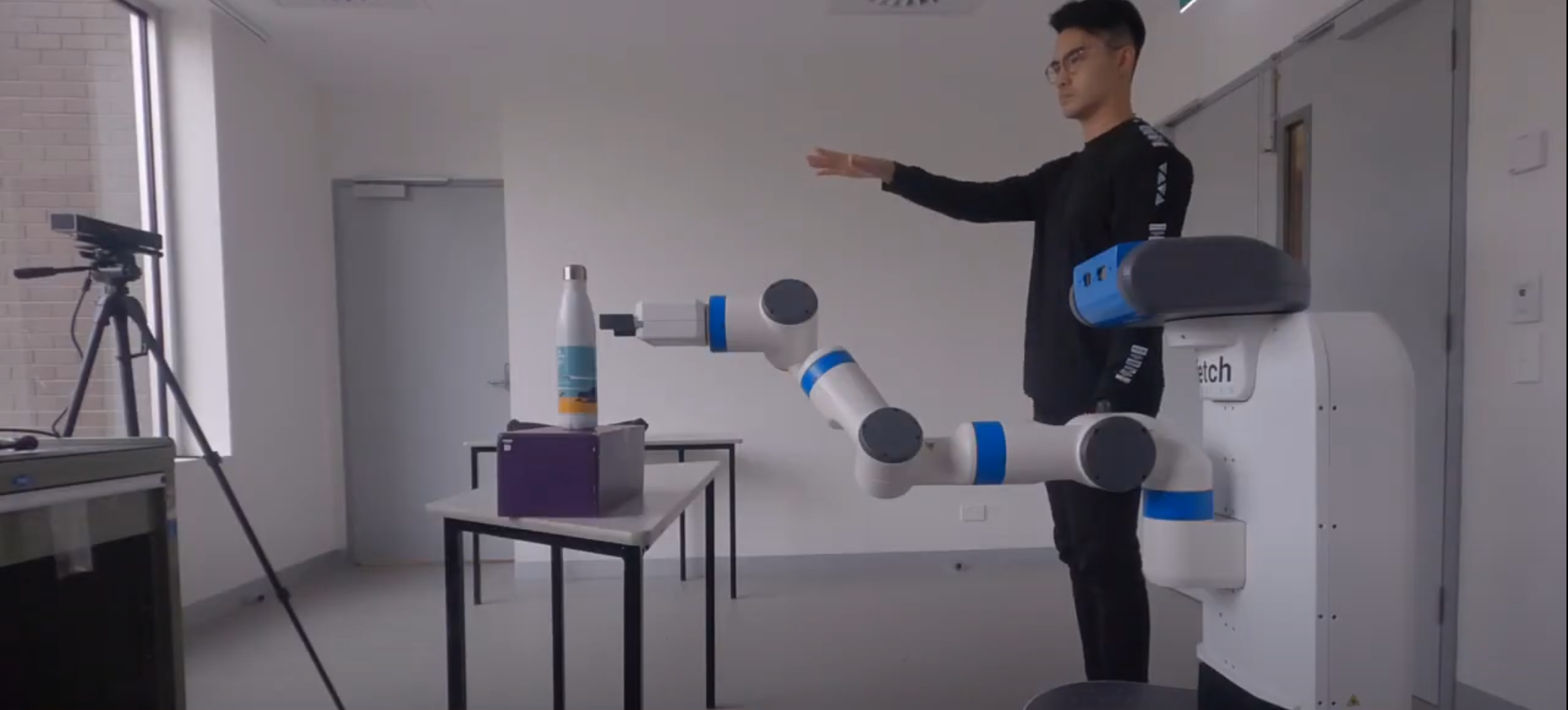}
    \vspace{-5mm}
    \caption{Demonstration of the gesture recognition system}
    \label{fig:setup}
    \vspace{-6mm}
\end{figure}


The goals of this paper are 1) to examine if different people inherently use similar gestures to communicate a given message or command to a robot and 2) to build a reusable gesture recognition system for humans to communicate with different robots in HRI applications. Our main contributions are as follows:
(1) We propose a set of commands useful for human-robot communication during interaction.
(2) We conducted a user study to investigate how participants would gesture to communicate this set of commands to a robot, and collected a set of uninstructed gestures across individuals. Large inter-person variations were found from the collected data, indicating the need for a standardised set of gestures.
(3) We proposed an initial standardised set of communicative hand and body gestures for the list of common commands in HRI tasks.
(4) We trained a robust gesture recognition model for the proposed gestures using a relatively small amount of training data.
(5) We integrated the resulting gesture recognition model into the ROS framework, opening up the opportunity for future ROS applications.

The structure of this paper is organised as follows: Section II discusses the related works, with the proposed method for training the gesture recognition model in Section III. The results are shown in Section IV, accompanied by a demonstration in Section V. Discussions and conclusion are presented in Section VII and VIII.

\section{RELATED WORK}
        \label{sec:related_works}
        \subsection{Communication in Human Robot Interaction}


Current HRI research focuses on improving the intuitive nature of gestural communication, to facilitate close cooperation between humans and robots~\cite{6483609}. There is a need to establish a communication protocol to facilitate proximate interaction between humans and robots~\cite{noauthor_1_nodate}. Existing research has investigated interactive channels such as voice, speech, and gestures~\cite{6926324, doi:10.1089/dst.2012.0010}. Speech is a common method for human communication, and a voice-based interaction is often proposed to facilitate a more natural, collaborative human-robot synergy~\cite{10.1145/3171221.3171280, 6926324, abdelhamid_roboasr_2012}. Automatic speech recognition can allow robots to understand user commands without requiring specialised user training~\cite{matuszek2014learning, 10.1145/3171221.3171280}. However, current speech-based applications can experience challenges, such as the complexity of accents, dialects, ambiguity in context, grammar irregularities, homophones, and operation performance in noisy environments~\cite{doi:10.1089/dst.2012.0010}. Instead, gestures can provide a non-verbal vision-based form of communication~\cite{6144164, 8260797} that can be used in noisy environments and with cost-effective sensors (i.e., digital cameras)~\cite{6144164, 1613091}. The challenge is to operate these systems in complex scenes with different backgrounds and variable lightning conditions, taking account different gesture positions, orientation and occlusions~\cite{6144164, 5356486}.

\begin{figure*} [t]
    \centering
    \includegraphics[width=0.98\textwidth]{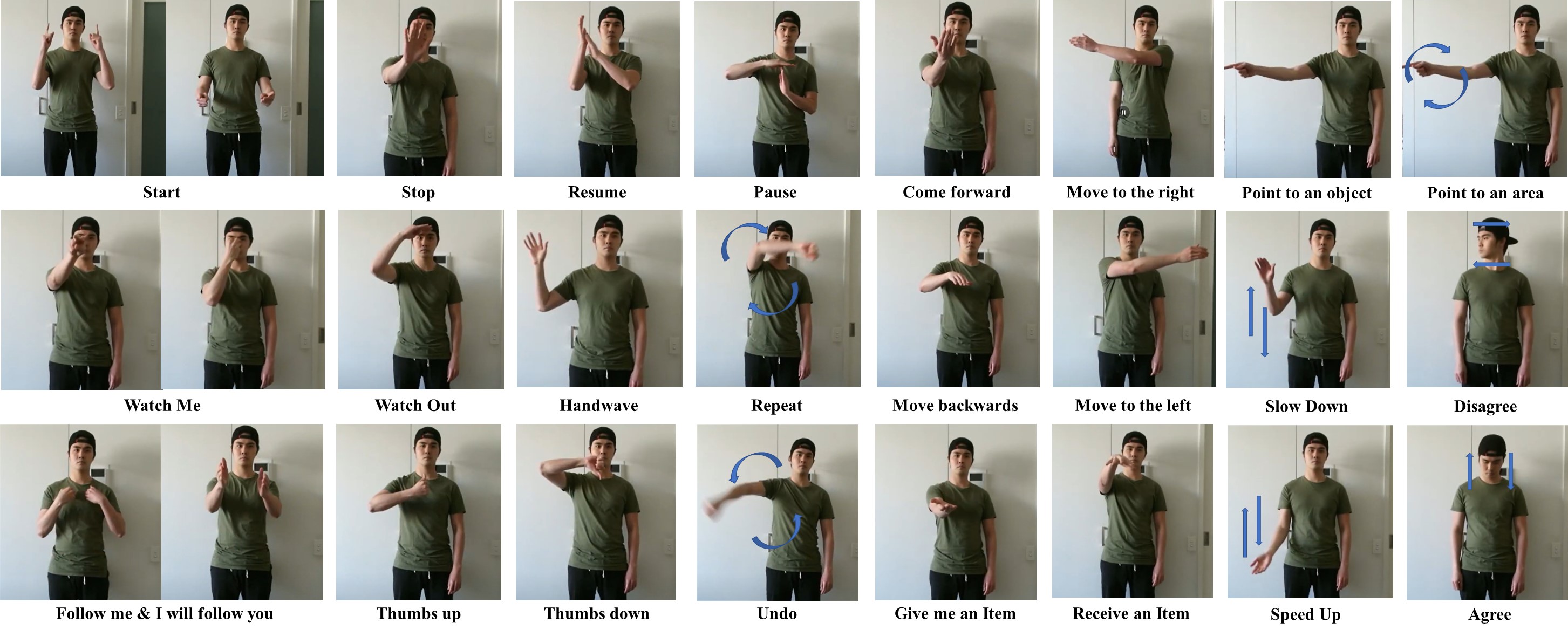}
    \vspace{-3mm}
    \caption{Snapshot of the proposed set of communicative gestures for Table~\ref{table:commands}. Video demonstrations are available in the project repository. \protect\footnotemark}
    \vspace{-5mm}
    \label{fig:gestures}
\end{figure*}
\footnotetext{Project repository: https://github.com/alberttjc/hiROS}

\subsection{Gesture Recognition Models}
Different approaches have been adopted to develop a real time gesture recognition model. The following section will focus on works that are capable of real time recognition using a single RGB camera.

One popular approach is using 2D or 3D key human joint positions detected by skeleton trackers like OpenPose~\cite{cao_realtime_2017}. This approach provides the developed gesture recognizer with the capability of detecting multiple persons and their corresponding gestures within an image frame. While effective, the reduced spatial information generated from 2D poses has been shown to limit its accuracy~\cite{luvizon_multi-task_2020,angelini_actionxpose_2018}. To achieve better accuracy, the use of multiple cameras to obtain 3D poses was also investigated but the requirement of a multi-camera setup makes such systems less suitable for stand-alone robot applications~\cite{pham_unified_2020}.

Another popular approach uses 2D or 3D Convolutional Neural Networks (3D CNN) to perform human action recognition from video sequences. These networks use spatio-temporal filters as the main building block, with the fully-connected features of the network transferred into a recurrent neural network to encode long-term temporal information~\cite{neurons_gesture_2017, materzynska_jester_2019, Zhang_2017_ICCV}. The resulting model is able to achieve accurate human activity recognition in real time, with the performance attributed to a large database used for training~\cite{8922847}. To overcome the lack of training data, transfer learning can be investigated, building a custom classifier using a pre-trained model and a small amount of new additional training data.

\subsection{Existing Gesture Datasets}

Existing human activity datasets often focus on human actions, covering sport actions, hand gestures, daily activities and training exercises~\cite{chen_utd-mhad_2015, noauthor_berkeley_nodate, peng2017twostream, Wang_2015, Damen2018EPICKITCHENS, xiao2018action}. However, the majority of the proposed gestures are not communicative in nature. In addition, the data sets were often not released for use, which made replication more difficult. Other datasets included gestures performed by a small number of participants in carefully controlled settings. This can result in poor generalization when a model trained on such datasets is deployed in varied settings or user demographics~\cite{5995347, gong2013reshaping}.

Such concerns can be addressed using the 20BN-jester dataset, a large-scale dataset consisting of 148,092 short, densely labeled video clips of humans performing generic hand gestures in an unconstrained environment~\cite{materzynska_jester_2019}. Participants performed one of 25 gestures in front of a camera, capturing their upper body view. This is incompatible for HRI applications as humans do not commonly interact with robots in such proximity. Additionally, not all gestures are communicative and may not be applicable for HRI. Hence, we decided to develop our own dataset by innovating and expanding upon this dataset.



\subsection{Gestural communication for HRI}

Using vision-based gestural communication, humans can communicate with a robot and replan interaction tasks in real time ~\cite{chen_online_2019}. Existing work has explored gestural communication in proximate interaction, using various forms of gestures to send commands to control a robot. Communication with humanoid robots, industrial robots and unmanned aerial vehicles has been explored, in order to improve the social interaction between the human and robot or to execute user-specified tasks ~\cite{chen_online_2019, kim_vision-based_2013, canal_gesture_2015}. The gesture sets used in these works are typically unique to each robot application, often self-defined or referenced from literature ~\cite{6739488,article,9427388,https://doi.org/10.1111/exsy.12490}. These gesture sets and associated gesture recognisers are not easily generalised for other HRI applications.

\section{METHOD}
        \label{sec:method}
        In the following, we first introduce the gesture recognition model chosen for this paper, followed by proposing a set of commands intended for common HRI applications. Next, we detail the data collection process conducted in our user study. The data collected from the user study is then used to train a custom gesture recognition model, to implement the proposed gestures in different robot applications.

\subsection{Gesture Recognition Model}


The gesture recognition model presented in~\cite{neurons_gesture_2017} was chosen due to its high accuracy and reliability in recognising dynamic human gestures in real-time, with the ability to train a custom recognition model using a relatively small amount of training data. The model utilises two 3D Convolutional Neural Networks (CNN) to extract spatiotemporal features, a LSTM (long short-term memory) layer to model longer temporal relations, and a softmax layer that outputs class probabilities. In contrast to 2D CNNs, which are good at image processing, 3D CNNs use three-dimensional filters which extend the two-dimensional convolutions into the time domain. Using such 3D filters in the lower layers of a neural net is used to consider both spatial and temporal features of gestures.

The selected model was pre-trained with~\cite{materzynska_jester_2019}, a large dataset of short, densely labeled video clips of human actors performing generic hand gestures in front of a webcam. As each video clip from this dataset captures different background, such as sub-optimal lighting conditions and background noise, training with this dataset forces the neural network to learn the relevant hierarchy of visual features that can separate signal (human motion) from the noise (background motion)~\cite{neurons_gesture_2017}. As the intended use case of this model is for fitness tracking, the gestures included may not be suitable for human-robot communication. However, we can leverage the pre-trained weights from this model to train a custom recognition model and adapt it into the ROS framework.

\subsection{A Proposed Set of HRI Commands}
To develop the communicative gesture set, we first generated an initial set of commands intended for use during human robot interaction, such as commanding common actions to a robot (Table \ref{table:commands}). We then proposed a gesture for each command (Figure \ref{fig:gestures}), taking consideration of different criteria to promote functionality and usability in both social and research capacities. This protocol was designed by further developing or innovating upon gestures from existing datasets~\cite{materzynska_jester_2019, chen_utd-mhad_2015}, including exploration of literature on communicative gestures from other domains, such as the Australian Sign Language (AusLAN), military gestures and scuba diving hand signals ~\cite{Auslan,14950e1796664051a70c0b71940b9c3c,behnke_scuba_2015}. The proposed gesture set focuses on facilitating collaboration for Human-Robot Interaction, novel to existing datasets~\cite{materzynska_jester_2019, chen_utd-mhad_2015, xiao2018action}. Referring to Barattini et al., gestures should be distinguishable, natural, socially acceptable, and easy for people to remember \cite{barattini_proposed_2012}.

To ensure these gesture criteria are met, detailed evaluation and testing from the proposed set of gestures was collected, internally and externally from the research team. Gestures that do not meet the chosen criteria were then revised to ensure the final proposed set meets the criteria.  

As a result, a total of 25 gestures were proposed with Figure~\ref{fig:gestures} illustrating a brief snapshot. From~\cite{materzynska_jester_2019}, two additional gestures of "doing nothing" and "doing something else" were also introduced. These two classes are used to identify when the user is idle or performing  non-communicative gestures.

\begin{table}[b]
    \vspace{-3mm}
    \caption{Proposed list of commands for HRI applications}
    \centering
    \label{table:commands}
    \begin{tabular}{l l l}
            1. Start         & 10. Point to an Object    & 18. Thumbs up          \\ 
            2. Stop          & 11. Point to an Area      & 19. Thumbs down        \\
            3. Handwave      & 12. I will Follow You     & 20. Give me an item    \\
            4. Resume        & 13. Follow Me             & 21. Receive an item    \\ 
            5. Pause         & 14. Watch Me              & 22. Move backwards     \\ 
            6. Agree         & 15. Watch Out             & 23. Come forward       \\ 
            7. Disagree      & 16. Speed up              & 24. Move to the left   \\
            8. Repeat        & 17. Slow down             & 25. Move to the right  \\
            9. Undo          &                           &                        \\ 
    \end{tabular}
\end{table}

\subsection{Data Collection Study}
We conducted a data collection study with two main objectives. The first objective is to determine whether different people would use similar gestures to convey the same commands when communicating with a robot. The second objective is to collect a set of demonstrations of communicative gestures, that can be used to train our gesture recognition model. The study was approved by the Monash University Human Research Ethics Committee (Project ID: 28756).

In the user study, participants were asked to consent to data collection, provide demographic data, rate their personal experience with technology, along with their preference of interacting with a robot in a domestic setting. The data collection process consisted of 2 stages. In Stage 1, participants were asked how they would gesture to a robot to best convey a given intention/command (e.g., 'How would you gesture to tell a robot to stop in place?'). This stage was intended to collect more information on what gestures participants would naturally choose to use if they were given no prompts or video demonstrations. In Stage 2, participants were shown a video demonstration of how to gesture each command, and then asked to perform each demonstrated gesture after viewing the video example. A total of 27 videos were presented in a randomized order. These videos showcase the proposed gestures from Figure~\ref{fig:gestures}, which were performed and recorded by a research team member.
To collect a diversified dataset, Amazon Mechanical Turk (AMT) was also used to collect demonstrations from people from different geographical locations. To ensure data quality, video entries from each participant were manually reviewed to exclude any invalid submissions (e.g., participant is not visible or not performing any requested gestures). Participants who passed the quality assurance check were compensated with \$2 USD.

Before collecting any data, participants were asked to ensure the web application was functional, with the camera positioned at eye level where it can capture the full or half upper body view of the participant.Participants were asked to conduct the study in a comfortable area, with sufficient lighting and space. Appropriate and comfortable attire was recommended, with the option of using a face mask if preferred. Participants were asked to perform each gesture with the intention to communicate with a robot, with no others present in any of the recordings. These instructions were listed in the study, along with an image to reference the proposed setup.

For data collection, a website was created and hosted online with the cross-platform compatibility to conduct the user study worldwide. Data from the user study was stored in a secure facility hosted by Monash University. With the COVID-19 pandemic, participants were not allowed in the laboratory to conduct this study with a robot. Instead, participants were given a reference picture and detailed instructions through the study. 


\section{RESULTS}
        \label{sec:results}
        In the following, we describe our findings from the data collected from the user study. We then trained the gesture recognition model with the collected data, using a k-fold cross-validation scheme. Datasets collected from each stage were split into 5 folds, with one set aside for testing and the remainder used for training. This process is repeated 5 times, with each fold used as test data. The overall results were pooled together over the 5 folds for evaluation. Each iteration was trained for 100 epochs, with the model converging after epoch 80.

\begin{figure}[t]
    \centering
    \includegraphics[width=.4\textwidth]{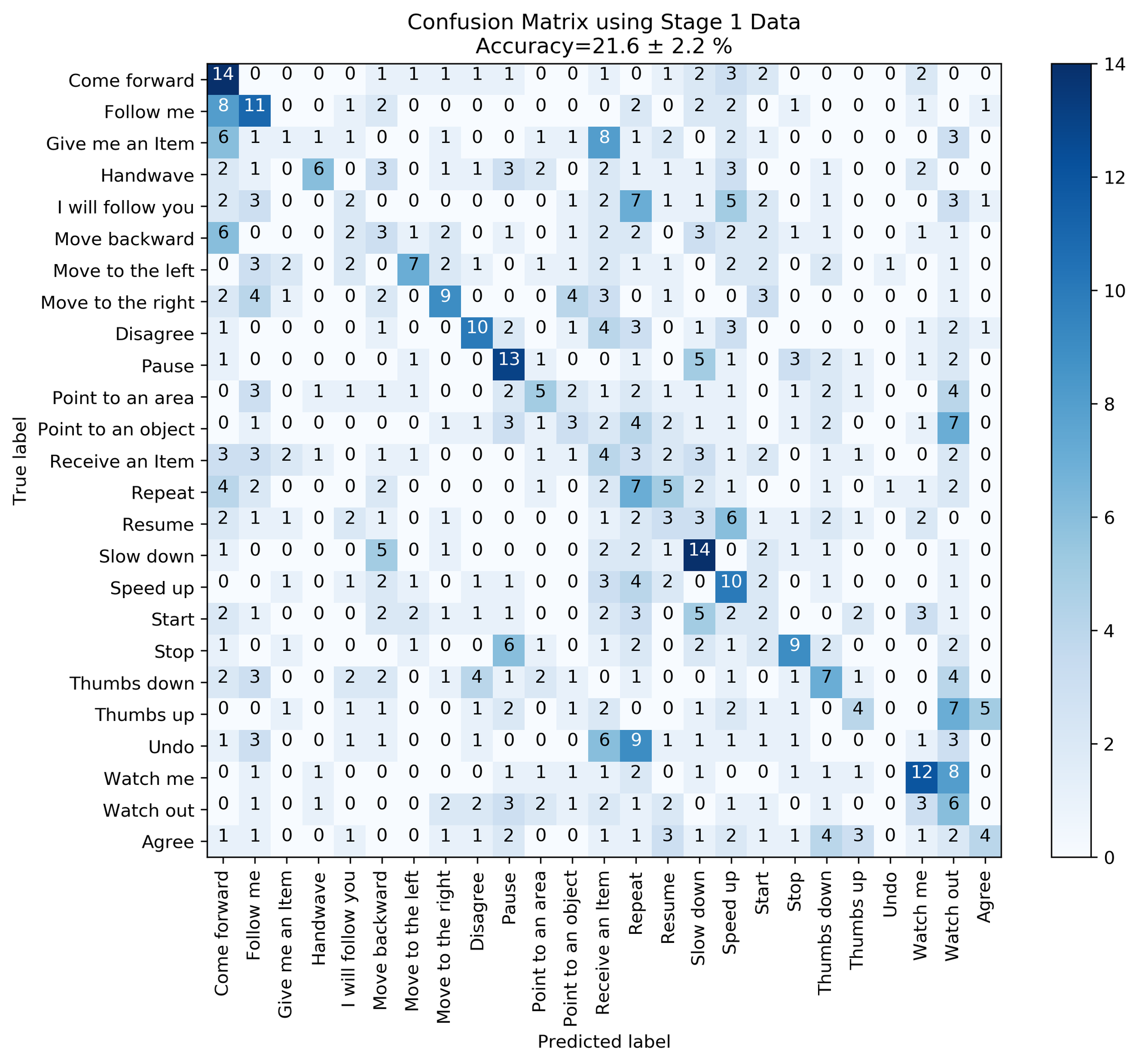}
    \vspace{-3mm}
    \caption{Confusion matrix generated using training data from Stage 1 of the collection study - gestures curated from participants}
    \label{fig:stage1}
    \vspace{-2mm}
\end{figure}

\begin{figure}[t]
    \centering
    \includegraphics[width=.4\textwidth]{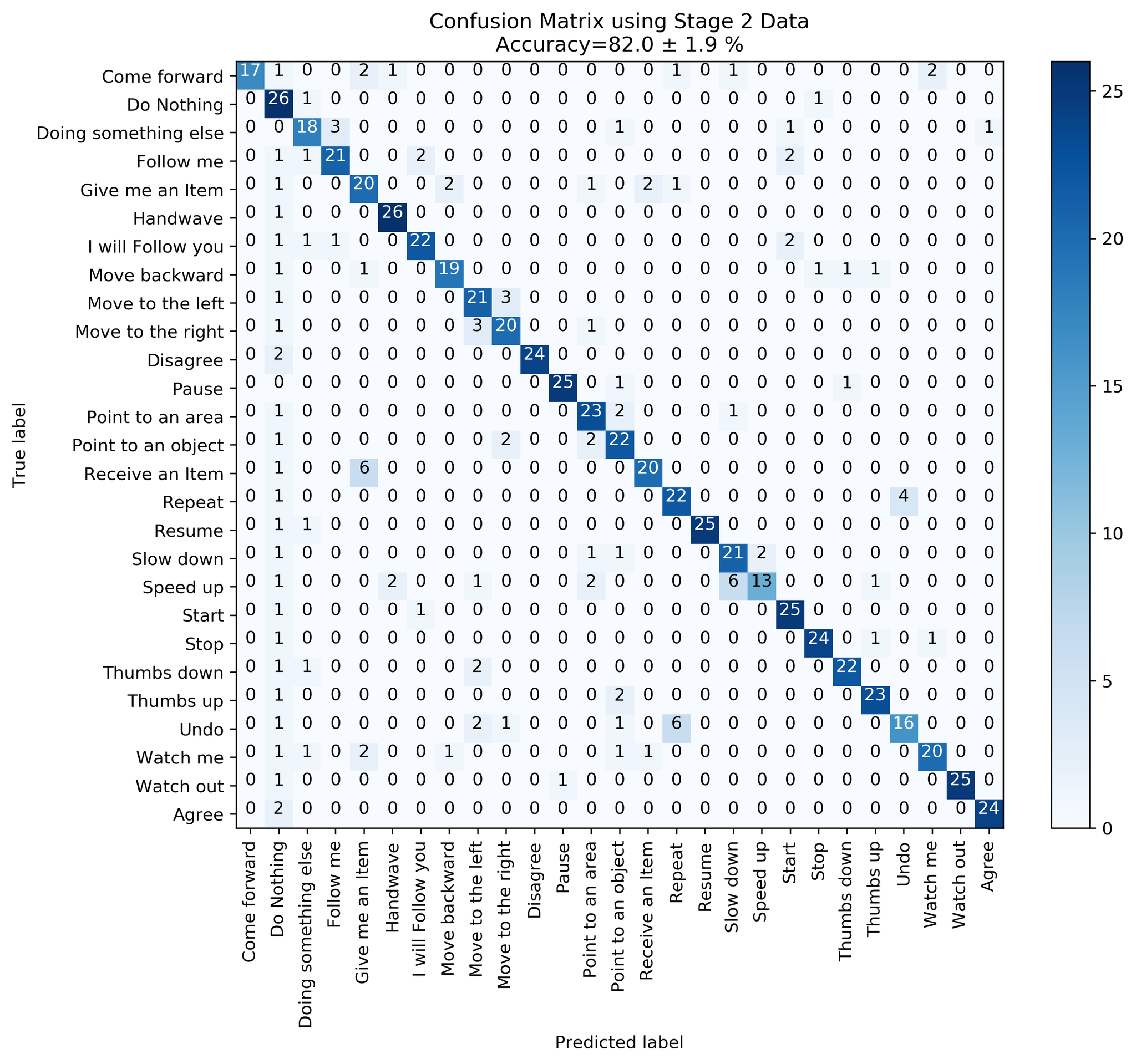}
    \vspace{-3mm}
    \caption{Confusion matrix generated using training data from Stage 2 of the collection study - demonstrated gestures}
    \label{fig:stage2}
    \vspace{-5mm}
\end{figure}

\subsection{Participant Sample}

A total of 190 participants took part in the user study, with 17 (9\%) recruited from the general public and 173 (91\%) from AMT. However, only 135 (71\%) participants completed the full user study by providing demographic data and video demonstrations for all gestures, while the remainder (29\%) provided only partial video demonstrations or demographic data. 74\% ($n=100$) of the participants identified themselves as men and 24\% ($n=33$) as women, with the remainder as "other" ($n=1, 1\%$) and "prefer not to say" ($n=1, 1\%$). The mean participant age and height was 31 years old (min. 19, max. 62) and 170 cm tall (min. 149cm, max. 192cm), with 90.4\% ($n=122$) being right hand dominant, 8.2\% ($n=11$) being left hand dominant and 1.4\% ($n=2$) being ambidextrous. 73.3\% ($n=99$) of participants used a built-in webcam to conduct this study, with 17.8\% ($n=24$) using an external webcam and 8.9\% ($n=12$) using their phone. A total of 60.5\% ($n=115$) participants agreed to release their video gestures to an open-access database. 

71.0\% of the participants ($n=135$) had no prior experience in the robotics field ($\Bar{x}=1.33,\sigma=0.56$) or had barely interacted with a robot ($\Bar{x}=1.24,\sigma=0.50$). They were also inexperienced in the artificial intelligence field ($\Bar{x}=1.35,\sigma=0.59$) and were unfamiliar with the data collection process for such projects ($\Bar{x}=1.42,\sigma=0.65$). Participants were familiar with using gestures to communicate with another person ($\Bar{x}=7.28,\sigma=1.82$), and showed no preference between using gestures ($\Bar{x}=1.46,\sigma=0.64$), verbal commands ($\Bar{x}=1.76,\sigma=0.67$) or both ($\Bar{x}=1.91,\sigma=0.78$) to interact with a robot in a domestic setting. This also extends to their preference between using a computer or laptop terminal ($\Bar{x}=1.61,\sigma=0.70$) and a handheld device ($\Bar{x}=1.79,\sigma=0.74$).

\begin{table}[b]
    \vspace{-3mm}
    \caption{Distribution of participants based on country of birth and country they spend most of their time in, grouped by continent}
    \label{table:continent}
    \centering
    \begin{tabular}{|c|c|c|}
        \hline
            Continent & Country of  & Country participants spend \\ 
                      &   Birth     & most their time in \\ \hline
            Africa          & 5.9\%     & 5.9\%     \\ 
            Asia            & 37.0\%    & 31.9\%    \\
            Europe          & 17.0\%    & 20.7\%    \\ 
            South America   & 11.9\%    & 12.6\%    \\
            North America   & 17.8\%    & 19.3\%    \\ 
            Oceania         & 10.4\%    & 9.6\%     \\
        \hline
    \end{tabular}
\end{table}

Information on the country where participant spend their most time in and their country of birth were collected. This distribution was grouped by continent, summarised in Table~\ref{table:continent}.

\subsection{Gesture Recognition Model}

\begin{table}[b]
    \caption{Model accuracy with different dataset size}
    \centering
    \label{table:performance}
    \begin{tabular}{|c|c|c|}
        \hline
            Average number of videos & \multicolumn{2}{ c |}{Accuracy} \\
            \cline{2-3} per gesture & Stage 1 Dataset & Stage 2 Dataset  \\ \hline
            $\sim50$  & $17.1\pm{3.3}\%$    & $81.0\pm{1.3}\%$  \\ 
            $\sim100$ & $19.7\pm{2.6}\%$    & $79.2\pm{2.1}\%$  \\ 
            $\sim150$ & $21.2\pm{2.2}\%$    & $82.3\pm{1.9}\%$  \\ 
        \hline
    \end{tabular}
\end{table}

We trained two separate gesture recognition models with the data collected from Stage 1 and Stage 2. The resulting confusion matrices are illustrated in Figure~\ref{fig:stage1} and~\ref{fig:stage2}. To gain further insight, we trained the gesture recognition model for each stage using different datatset sizes and compared the resulting performance (Table~\ref{table:performance}). Inspecting Table~\ref{table:performance} with the Stage 1 dataset, the accuracy of the model improved minimally despite using a larger dataset, while the model using Stage 2 data was more consistent with much better accuracy. As both models were trained using the same network architecture with a relatively small dataset, the results suggest that there is more variation in the data from Stage 1, making it more difficult for the model to achieve good performance. This supports the claim that different people may use different gestures to convey the same message. 

Comparing the confusion matrices shown in Figure~\ref{fig:stage1} and~\ref{fig:stage2}, we see that the classification results are much more concentrated on the diagonal in Figure~\ref{fig:stage2}, indicating a much higher accuracy and lower misclassification rate of the gesture recognition model trained using Stage 2 data, compared to that using Stage 1 data. In Stage 1, we asked the participants to use gestures that they think are most intuitive. The poor accuracy of the resulting gesture recognition model is most likely due to large variations in the dataset, meaning that different people do not inherently share a common set of gestures for conveying the same message. This is an important finding, as this may assist in explaining higher errors or incorrect classification often seen in other studies when people are not provided with instructions or demonstration on gesture patterns for robot interaction. 



Further examining the resulting confusion matrix from the gesture recognition model trained using our proposed set of gestures (Figure~\ref{fig:gestures}), we noticed that there is a small number of gestures that performed poorly within the inference layer. Distinctive and unique gestures such as "Handwave" and "Pause" performed well consistently, while gestures such as "Give an item" and "Come Forward" performed poorly. Referring to Figure~\ref{fig:stage1} and~\ref{fig:stage2}, gestures that performed poorly were shown to have a recall score lower than 0.85, regardless of their precision score. For example, "Give an Item" and "Receive an Item" (Table~\ref{table:commands}) both performed poorly and have a recall score of 0.66 and 0.71, respectively. Thus, in our final implemented gesture recognition system (demonstration with the Fetch robot in Section~\ref{sec:demonstration}), we removed the following ("Receive an Item", "Give an Item", "Speed up", "Slow down" and "I will follow you") to decrease the probability of error. Our final model was then able to achieve an accuracy of $84.1\pm{2.4}\%$.

\section{DEMONSTRATION}
        \label{sec:demonstration}
        \begin{figure} [b]
    \label{fig:system} 
    \centering
    \includegraphics[width=0.45\textwidth]{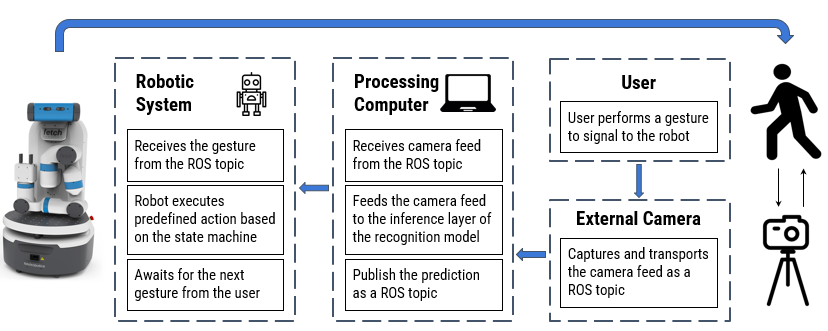}
    \vspace{-2mm}
    \caption{Overview of the implemented gesture recognition system}
    \vspace{-4mm}
\end{figure}

We integrated the trained recognition model into the ROS framework to provide an end-to-end system demonstration for robot application. Figure~\ref{fig:setup} depicts the interaction between the components of our implemented system. While a camera on the robot itself can be used, we demonstrate the possibility of using an external camera. The camera is placed in front of the user where it can capture either full or upper body view, with the data stream fed into the processing computer running the gesture recognition model.

In the demonstration, a human can interact with a Fetch robot through the proposed set of gestures (Figure~\ref{fig:gestures}). The interaction begins with a human sending a command (Table~\ref{table:commands}) to the robot by performing the proposed gesture (Figure~\ref{fig:gestures}). The gesture recognition model processes the video stream as a ROS image topic and publishes the model prediction as a ROS message. A state machine processes these results and instructs the Fetch robot to execute the corresponding action. The Fetch then remains idle until it receives the next command from the human.

In our demonstration \footnote{Demonstration video: https://youtu.be/X1rc98xzlJw}, we successfully sent commands to instruct a mobile manipulator robot to execute arm and base movements, as well as an object grasping and handover task. We used the gesture "start" to signal the robot base and directional gestures, such as "come forward" and "move to the right", to navigate to the intended direction. To move the arm, we use "point to an item" and directional gestures, followed by "resume" to perform grasping and handover task. To reset the position of the arm, the gesture "undo" was used. Moreover, we were able to send commands to pause, resume, or shutdown the gesture recognition interface when desired. This was done using the gestures "pause", "resume" and "stop", to specify to the robot when to pay attention to the gestures. While the demonstration was successful, the remote communication between the processing computer and Fetch robot experienced minor delays. To minimise these delays, we ensured the connected network had high bandwidth and asserted short delays between each prediction result, to avoid message queue overflow.

We implemented this system using the ROS framework to allow easy reuse and integration by other robot applications, with the source code available online\footnote{Project repository: https://github.com/alberttjc/hiROS} for researchers to adapt to different use case. A future user study will further substantiate the utility and evaluation of this system once the COVID-19 pandemic has been resolved to allow studies to recommence.

\section{DISCUSSION}
        \label{sec:discussion}
        Inspecting the gesture demonstration videos provided by participants in Stage 1, we found that there is a large variation in the gestures used across individuals for conveying the same given message. Using Stage 1 data, the resulting confusion matrix (Figure~\ref{fig:stage1}) had a high number of false positives. Using our proposed set of communicative gestures to guide a more constrained dataset in Stage 2, it was shown that a gesture recognition model of much higher accuracy can be obtained (Figure~\ref{fig:stage2}, Table~\ref{table:performance}). We believe a similar performance can be achieved in Stage 1 by collecting a much larger dataset, capturing the large variations in inherent gestures used by different people. However, by using a standardized set of gestures, we can train a good model using a smaller amount of training data, making it easier to incorporate new gestures or modify existing gestures when customizing for different applications. 

The trained model (from Stage 2) was able to achieve a relatively high accuracy, but several gestures were found to perform poorly compared to others. Those gestures included "Give an Item", "Receive an Item". This was likely the result of different gestures sharing similar motion and mannerisms, such as "Stop" and "Give an Item". Additionally, some gestures may be less effective when performed using different hands. As 90.4\% ($n=122$) of the participants were right hand dominant, gestures performed using the left hand may lead to varying results. This is a current limitation of our trained model, which we aim to address in future work. 


While using a standardised gesture set improves recognition performance, the gestures introduced in a standardised gesture set are limited and may not incorporate all gestures needed for different use cases. Users will also need to be trained on the proposed gestures, potentially resulting in less intuitive communication. To address these concerns, we have given consideration for a number of criteria while designing our gestures, aiming to make these gestures intuitive and easy to learn. Also, results in Table~\ref{table:performance} showed that a gesture recognizer with good accuracy can be achieved with as few as $\sim50$ demonstrations per gesture. This allows other users to easily add new gestures, to adapt for their own applications.

The current system does have strengths and limitations. The system is unable to accurately recognise gestures in the presence of multiple users, likely due to the fact that we instructed participants from the data collection study to perform these recordings alone. To address this, a human detection module could be integrated into the system to identify regions of interest, and enable gesture recognition of multiple users in the future. Using the selected neural network architecture and the pre-trained model, a relatively small amount of training data was needed to develop a robust gesture recognition model for our physical robot system demonstration. The current system allows users to interact with robots using the proposed communicative gestures, sending commands to perform simple tasks. The system is implemented in the widely used ROS platform for further adaption and use on different robot systems, with the code and a subset of data open-sourced. This will allow future research and development to more easily integrate and build upon our work for other applications.

\section{SUMMARY AND CONCLUSION}
        \label{sec:conclusion}
        In this paper, we have proposed a set of communicative gestures to provide a  foundation for future works in human-robot interaction via gesture recognition. We hypothesized that a standardised set of gestures is needed as we believed that there would be a large variation of gestures used across individuals to convey the same message. To validate our hypothesis, we conducted a data collection study and had participants perform gestures in two stages; Stage 1 where they were asked to come up with gestures they believe that best conveys the given command, and Stage 2 where they were shown and asked to reproduce the proposed gestures for each command. Results confirmed that there were large variations in gestures collected in Stage 1, hence, supporting our hypothesis. Using data collected in Stage 2, a gesture recognition model was then trained, and a gesture recognition system for a robot application was developed and demonstrated using the ROS framework. The developed system allows users to communicate with the robot using gestures and send commands to the robot to perform simple tasks. The current model demonstrates a good foundation for real time gesture recognition for HRI applications. The code and a subset of data for the developed system is available open-source, aiming to allow other researchers to easily integrate and extend for other applications. This includes the use of the ROS framework to allow for easy adaptation of the system for other robot types and resulting robot actions.

\bibliographystyle{IEEEtran}
\bibliography{ref}

\end{document}